\documentclass{article}
\usepackage{graphicx}

\usepackage{booktabs}   
\usepackage{subcaption} 
\usepackage[colorinlistoftodos]{todonotes}

\begin{document}

\title{Unifying AI Algorithms with Probabilistic Programming using Implicitly Defined Representations}       


\author{Avi Pfeffer \\ Charles River Analytics \\ apfeffer@cra.com
\and Michael Harradon \\ Charles River Analytics \\ mharradon@cra.com
\and Joseph Campolongo \\ Charles River Analytics \\ jcampolongo@cra.com
\and Sanja Cvijic \\ Charles River Analytics \\ scvijic@cra.com}

\date{}

\maketitle

\begin{abstract}
We introduce Scruff, a new framework for developing AI systems using probabilistic programming. Scruff enables a variety of representations to be included, such as code with stochastic choices, neural networks, differential equations, and constraint systems. These representations are defined implicitly using a set of standardized operations that can be performed on them. General-purpose algorithms are then implemented using these operations, enabling generalization across different representations. Zero, one, or more operation implementations can be provided for any given representation, giving algorithms the flexibility to use the most appropriate available implementations for their purposes and enabling representations to be used in ways that suit their capabilities. In this paper, we explain the general approach of implicitly defined representations and provide a variety of examples of representations at varying degrees of abstraction. We also show how a relatively small set of operations can serve to unify a variety of AI algorithms. Finally, we discuss how algorithms can use policies to choose which operation implementations to use during execution.
\end{abstract}

\section{Introduction}
\label{sec:introduction}
\newcommand{\sfunc}{\texttt{SFunc}}
\newcommand{\sfuncs}{\texttt{SFuncs}}
\newcommand{\opimpl}{\texttt{OpImpl}}
\newcommand{\opimpls}{\texttt{OpImpls}}
\newcommand{\opperf}{\texttt{OpPerf}}
\newcommand{\opperfs}{\texttt{OpPerfs}}
Probabilistic programming (PP) has made significant progress over the years. Early papers~\cite{koller1997effective, sato1997prism} contained very limited languages with one-off algorithms and were focused more on theory than practical applications.
As the first generation of practical systems developed (e.g. ProbLog~\cite{de2007problog}, IBAL~\cite{pfeffer2007design}, Infer.NET~\cite{winn2009probabilistic}, Church~\cite{goodman2012church}, Stan~\cite{carpenter2017stan}), the focus was on greater practicality of representation and efficiency of algorithms. Some languages, like Hakaru~\cite{narayanan2016probabilistic}, attempt to establish fundamental principles of computation in probabilistic programming. Recent languages like Venture~\cite{mansinghka2014venture} and Gen~\cite{cusumano2019gen} have focused on frameworks for organizing inference, while other languages like Edward~\cite{tran2016edward} and Pyro~\cite{bingham2019pyro} offer increased scalability and development of practical applications. Our goal is to continue this tradition and develop PP as a major organizing framework for development of intelligent systems.

In recent years, we proposed Scruff (citation withheld for anonymity) to be a framework for building AI systems based on predictive processing. As we have developed Scruff, we have come to realize that achieving our goal of developing a general AI framework requires a new way of writing PP systems. The challenges include:
\begin{itemize}
	\item If we are to develop a general framework for AI, components of a system take a variety of forms, such as code with stochastic choices, neural networks, differential equations, and constraint systems. Although only the first would traditionally be considered a probabilistic program, we would like to include them all under a single umbrella, with PP at its core as a Turing complete abstraction supporting native description of uncertainty.
	\item As the number of different kinds of model components increases, the difficulty of writing algorithms that work with all the different components also increases significantly. In particular, it is unlikely that any traditional algorithm (e.g. Monte Carlo sampling or belief propagation) will work with every kind of component.  It should be possible to use algorithms that are not fully general in applications that support them. 
	\item Furthermore, it may be that certain effects are achieved in different ways for different kinds of component. For example, if a distribution over the state of a variable at a particular point in time is required, this can be computed analytically for some ODEs, while others require simulation; for probabilistic models sampling can achieve the desired result, but again analytic methods could work for some models. Many PP languages only offer a single method of performing computations (e.g. Infer.NET uses expectation propagation); even those that offer a choice of algorithms typically require that once chosen, the same algorithm is applied to all model components (Stan?). It should be possible to combine different methods of computing within a single execution to best exploit sub-model properties.
\end{itemize}

Scruff's approach to addressing these challenges is centered on the idea of \emph{implicitly defined representations}. The idea is simple but has significant implications for the design of the system. We define a set of standard effects we might want to perform with model components and call them operations. These operations might include, for example, to generate a representative value for the component, to compute the probability of a value, to generate a representation of a probability distribution over the values of a component, to infer conditions on the inputs to a component given observations on its output, or to compute the gradient of parameters given observations of the inputs and outputs. 

The operations serve to separate the representation of individual model components from the implementation of algorithms. The term "implicit" comes from the fact that, as far as the algorithms are concerned, the components are defined implicitly by the operations they support; the specific representation of the components does not matter to the algorithms. Algorithms are written using these standardized operations and work for any components on which those operations are defined. For example, if an algorithm requires generating representative values for a component, it can use a \texttt{sample} operation that can be implemented for generative probabilistic models employing Monte Carlo sampling as well as ODEs employing simulation. These kinds of components can be freely combined in a model without any extra work. Furthermore, other modeling frameworks and algorithms can be wrapped with operations, enabling them to be incorporated into Scruff. These properties lead us to believe that Scruff does indeed have the potential to be a general organizing framework for AI systems. Furthermore, although the motivation has been PP, the ideas apply to any reasoning framework that involve connected model components. So, for example, we could potentially use Scruff to organize different kinds of deep learning components.

Implicitly defined representations enable Scruff to address all the challenges liste4d above. By making the implementation of operations the salient explicit property of a representation, algorithms can readily examine whether a particular model component supports a specific operation, and decide whether or not to use that operation. Meta-reasoners can also determine whether an algorithm can run on a model by examining whether the components support the operations required of the algorithm. In some cases, an algorithm might only require some of the components to implement an operation, while others do not. For example, for a feed-forward neural network (NN) component that processes an image to produce a classification, it is hard to compute a probability distribution over the image given the classification. An algorithm like belief propagation (BP) generally requires such a distribution, but for a leaf of the model, like an image classifier, it can be dispensed with. Combining NNs with BP is not new (e.g.~\cite{kuck2020belief, jia2021graph}); the innovation is that it is done in a fully general way and accomplished with little extra work in a common framework.

Scruff uses policies to control the invocation of operations. Policies stipulate which implementation of an operation to use or which values of hyperparameters should be chosen (e.g. number of samples in a histogram estimation). Policies can be specified using a small number of standard concepts (e.g. laziness), to provide a uniform way for operation implementers to communicate the choices and for algorithm designers to state their preferences. Policies can also be optimized to identify novel heterogenous operation execution patterns resulting in superior accuracy or speed.

From a design point of view, implicit operations have driven the designers of Scruff to push for increasing degrees of abstraction. The goal is to find a relatively small number of operations that support a large number of algorithms, while being sufficiently general to enable implementation on a wide variety of components. We have also sought to identify abstract representations of model components that include a large number of concrete implementations, enabling a relatively small number of operation implementations to hold for a wide variety of model components.

In this paper, we first present briefly the basic architecture of Scruff and terminology used, and give examples of some of the most frequently used operations.
We then present details of the implementation of the implicit operations framework, including policies.
We then describe some of the modeling abstractions used to support a large number of model components with a small number of operation implementations.
Then, we present examples of Scruff's support for different kinds of algorithms and choices.
Finally, we conclude with a discussion of future possibilities, including meta-inference and planning using operations.

\section{Scruff Architecture and Terminology}
\label{sec:scruff}
In the introduction, we used terms like "model" and "model component", which make intuitive sense, but Scruff has specific and precise terminology which we will use from now on.
Scruff is implemented in Julia and we will use Julia code to explain concepts.

One of the central concepts in Scruff is a \emph{stochastic function} (called \sfunc), which is a representation of a relationship from inputs to outputs, in which the same input can produce multiple outputs. We allow deterministic functions as a special case with a single output for each input. Although we use the term "stochastic", the probability distribution over outputs is not necessarily specified; operation implementations (\opimpls) that require a specific probability distribution (e.g. the closed-form expectation of a parameterized normal distribution) will not be supported by other \sfuncs, but it can still implement that operation in another way and participate in some algorithms. In addition, an \sfunc~is not necessarily represented by generative code as in most probabilistic programming languages, but can be any mathematical representation of the required relationship.

Formally, an \sfunc~is parameterized by three parameters, \texttt{SFunc\{I,O,P\}}. \texttt{I} is the type of inputs to the \sfunc, which must be a (possibly empty) tuple type. In other words, an \sfunc~has zero or more inputs, each of which can be any Julia type. \texttt{O} represents the output of the \sfunc, which can be any Julia type. In particular, \texttt{O}, as well as components of \texttt{I}, can be \sfuncs, which enables higher-order \sfuncs. Also, \texttt{O} can be the Julia type \texttt{Nothing}, which enables \sfuncs~to be used as entities that do not produce an output but score the inputs. \texttt{P} is the type of parameters to the \sfunc, which is important to machine learning with Scruff, but is not essential for this paper. Examples of \sfuncs~and their abstractions are presented in Section~\ref{sec:sfuncs}.

Scruff uses the term \emph{model} to describe something that generates an \sfunc~at different points in time to represent the evolution of some domain entity. Such a domain entity that evolves over time under a model is called a \emph{variable} in Scruff. We will use variable to denote an entity whose value is described by an \sfunc. While the model concept is central to Scruff's ability to reason about dynamically changing systems, they are not essential to the topic of this paper, and we will not use them further, other than to point out that we will avoid using the term model from now on because of its Scruff usage. We will restrict use of the term model to examples from the literature where the usage is standard.

For what we called "model" in the introduction, Scruff uses the term \emph{network}. A network consists of a directed acyclic graph over \sfuncs. Specifically, for each \sfunc~in the network, there is an assignment of an \sfunc~in the network to each of its inputs. Despite the acyclic structure of networks, the use of scoring \sfuncs~in which the output type is \texttt{Nothing} enables Scruff to represent undirected graphical models.

Recursively, a network is also an \sfunc. Some of the \sfuncs~without parents in the network are designated as placeholders, corresponding to inputs. Other \sfuncs~in the network are designated as outputs. The input of the network is then a tuple of output types of the placeholders, while the output of the network is a tuple of the outputs of the output~\sfuncs. As~\sfuncs, networks support operations that enable them to be used in algorithms.

An \emph{operation} is a representation of a type of computation performed on an \sfunc. It has a type signature that describes the expected input and output types, depending on the type parameters of the particular \sfunc. An operation also has a set of \emph{operation implementations}, or \opimpl, each of which is a Julia method that takes inputs and returns an output of the appropriate types for a given \sfunc. There may be more than one implementation of a particular operation for a given \sfunc~that employ different strategies. We defer more detailed discussion of operations to Section~\ref{sec:operations}.

An \emph{algorithm} is a program that uses operations on a set of \sfuncs~to perform some calculation.
An algorithm may use a \emph{policy} the defined how it dispatches against multiple \opimpls.
That policy can set the hyperparameters of an \opimpl,
for example expressing its preferences for runtime/accuracy tradeoffs.
BP~\cite{pearl1988probabilistic}, variable elimination (VE)~\cite{zhang1994simple, dechter1998bucket}, and sequential Monte Carlo~\cite{doucet2000sequential} are examples of algorithms in Scruff. 

Many of Scruff's algorithms (like sequential Monte Carlo) explicitly involve time and dynamics, but this is not a necessary feature of an algorithm. 
Just like a network can recursively be an \sfunc, so an algorithm that performs reasoning tasks on a network can become the implementation of an operation on the network as \sfunc~at the higher level. Scruff also includes a framework for runtime support of algorithms, providing services such as message passing and memory management.
Algorithms and policies will be discussed in Section~\ref{sec:algorithms}.

\section{Operations}
\label{sec:operations}

One of the primary contributions of Scruff is the design of an interface for computing and composing the application of higher-order functions on SFuncs. We refer to these higher-order functions as operations. Operations are associated with an intended mathematical function. For example, an \texttt{expectation} operation has the standard mathematical definition in a probabilistic framework of $\mbox{\texttt{Expectation}}_f(y \mid x) = \int y f(y|x)$.  
A particular \opimpl~may compute this statistic exactly, or it may use an approximate algorithm, possible controlled by a hyperparameter (e.g. number of samples).

Because the intended mathematical function is stated, algorithms generally know what to expect from implementations of operations.
Scruff does not have any means to verify the correctness of implementations, however. Instead, Scruff can verify that different operation implementations are compatible with one another to some degree of accuracy. For example, any \sfunc with vector output type that supports a \texttt{Sample} operation implicitly supports a generic \texttt{Expectation} operation via monte carlo method. Thus, if some closed form \opimpl for \texttt{Expectation} is implemented Scruff can verify the degree to which an existing sample-based implementation agrees with the new method.

Operations also define a type signature transformation that produces the type signature of the resulting operation from the type signature of the \texttt{\sfunc\{I,O,P\}}. For instance, an operation \texttt{SampleN} might produce a set of $n$ conditional samples from an \sfunc, where $n$ is an argument to the operation---in that case, we say that \texttt{SampleN} has signature \texttt{(I,O): (I, Int) -> Vector\{O\}}.
Any implementation of \texttt{SampleN} for a given \sfunc~must satisfy that defined signature.
Note that we use camel case for the operation itself, which is a type in Julia, and lower case for the \opimpl method, in accordance with Julia standards. In practice each `OpImpl` is assigned a unique name to enable controlled dispatch.

The creation of the appropriate type and function is achieved using a macro \verb:@op:.
Implementations of the operation for specific \sfuncs~are created using the \verb:@op_impl: macro.
Specifically, the type of the operation is parameterized by the type of the \sfunc~on which it is implemented.
The \sfunc~on which an operation is invoked is always the first argument to the operation's function; when the function is called, the \sfunc~is wrapped in the appropriate type to ensure that the correct \opimpl~is called.
Because the \sfunc~type is already present in the type of the \opimpl, it is omitted from the arguments in the \verb:@op_impl: definition.
For example, an implementation of \texttt{SampleN} for a \texttt{Flip} \sfunc~might be given by:

\begin{verbatim}
@op_impl function SampleN{Flip}(x::Nothing, n::Int) 
      # Flip has no parents, so x is Nothing
	  ps = rand(Float64, n)
	  # sf refers to the sfunc on which the implementation is defined
	  return [sf.prob < p for p in ps] 
end
\end{verbatim}

In contrast with typical method definition and usage, \sfuncs~may have multiple implementations of the same operation. This is common when there are fast, approximate ways to compute something as well as slow, exact approaches. Each \opimpl~can define a set of hyperparameters that affect the calculation performed. For instance, an iterative algorithm might specify the number of iterations, or a histogram calculation might internally take a number of samples. For convenience each hyperparameter must have an associated default value, though they can be changed manually or automatically at operation invocation. Consider an example of inverting a deterministic linear SFunc using a builtin procedure versus using an iterative algorithm. In the code below, the \texttt{struct} in each \texttt{@op\_impl} creates a specific type for the \opimpl.

\begin{verbatim}
@op_impl begin
    struct InvFullMatMul <: Invert{MatMul} end
    function invert(y)
        global matmul_invert_count += 1
    return sf.mat \ y
    end
end

@op_impl begin
    struct InvIterativeMatMul <: Invert{LinearOp}
        num_iters::Int = 5
        step_size::AbstractFloat = 1.
    end
    function invert(y)
        x = zeros(shape(sf))
        for i in 1:op_imp.num_iters
            res = sample(sf, x) - y
            x = x - op_imp.step_size*res
        end
    return x
    end
end
\end{verbatim}

Operations can take \sfuncs~as arguments and return them as well. This is critical for the operations used in algorithms like belief propagation.
Here we present simplified versions of definitions of the \texttt{ComputePi} operation, for \sfuncs~that only have a single parent, and
leaving out additional arguments that provide the ranges of the \sfunc~and its parent.
In this case, \texttt{ComputePi} takes a message from a parent, in the form of a \texttt{Dist}, which is an \sfunc~that defines an unconditional distribution, and returns a prior probability distribution over the child, also represented as a \texttt{Dist}.
In that section we also show a complete implementation of \texttt{ComputePi} that includes the additional arguments and works for any number of parents.

\begin{verbatim}
ComputePi: (I,O): Dist{I} -> Dist{O}
\end{verbatim}

In addition to specializing on the type of \sfunc~on which the operation is invoked, the operations can also specialize in their other arguments following the multiple dispatch design of Julia. For example, \texttt{ComputePi} for a normal distribution can be difficult and may require sampling from or discretizing the range of the normal distribution. When the parent message is normal, however, it can be computed in closed form. We can create a specialized implementation of \texttt{ComputePi} for this case. This example exploits the fact that \texttt{Normal} is a kind of \texttt{Dist}, so can be used as the parent message.

\begin{verbatim}
@op_impl function ComputePi{Normal}(ps::Normal)
    mu = expectation(ps)
    var = variance(ps) + sf.var
    return Normal(mu, var)
end
\end{verbatim}

\section{SFunc Abstractions}
\label{sec:sfuncs}
	
Since Scruff implementation is centered around the implementation of operations, it is vital to find the right abstractions that enable a small number of \opimpls~to apply to a large number of specific \sfuncs. These abstractions are centered around notions of composition, with the idea being to develop composition \sfuncs, in which the implementation of operations is derived automatically from the implementations of components. SFuncs can have input and output types described by other SFuncs, so it is possible to implement models described by expressive recursion - this ensures our SFunc composition representations are Turing complete.
	
\subsection{Basic abstractions}	
	
The most basic abstractions specialize on the types of inputs and outputs. We have already introduced \texttt{Dist}. In particular, we define
\begin{verbatim}
abstract type Dist{O,P} <: SFunc{Tuple{}, O, P}
\end{verbatim}	
to represent an SFunc with no parents (\texttt{Dist} stands for distribution for the most common case, although as before an explicit distribution does not need to be defined).
Examples of \texttt{Dist} include \texttt{Cat} for categorical, \texttt{Flip} for Bernoulli, and \texttt{Normal} \sfuncs. The \texttt{Dist} representation is useful because not having an input can make certain operations easier to define. For example, sending $\lambda$ messages for BP, an operation difficult to derive tractably from pure generative models, is trivial because there are no parents to send to.

Similarly, for an empty output type we define
\begin{verbatim}
abstract type Score{I,P} <: SFunc{I, Nothing, P}}
\end{verbatim}
Implementations of \texttt{Score} can be used to represent evidence or soft constraints, as well as implement undirected graphical models as discussed earlier.	Examples of \texttt{Score} include \texttt{HardScore}, which only allows one value; \texttt{SoftScore}, which weight different values; and \texttt{FunctionalScore}, in which the score is not represented explicitly but is defined by a function that takes a value and returns a score. This kind of functional score is useful when there are infinitely many values.
	
\subsection{Mixtures}	
	
A simple example of a composition \sfunc~is \texttt{Mixture}. Mixtures are a standard concept in probabilistic programming, in which the generative process randomly selects between one of several possibilities, each with a given probability. In a similar way, Scruff's \texttt{Mixture} randomly selects between several \sfuncs, but from the \texttt{Mixture}'s point of view the specific types of \sfuncs~is completely abstracted away; they could be probabilistic models, ODEs, NNs, or perhaps a mix of all these. The \opimpls~for \texttt{Mixture} invoke the corresponding operations on the components; the operation will be supported by the \texttt{Mixture} whenever it is supported by all the components, and the implementation will be derived automatically.

For example, here is the implementation of the \texttt{ComputePi} operation for \texttt{Mixture}. \texttt{ComputePi} is an operation used in BP, as defined by Pearl's classic algorithm~\cite{pearl1988probabilistic}. The purpose of this operation is to take a set of $\pi$ messages from the parents of the \sfunc, representing probability distributions over parent values, and compute the $\pi$ distribution for the \sfunc~itself, which is the expected distribution over values of the \sfunc~given the distributions over the parents. \texttt{ComputePi} takes three arguments: The range of the \sfunc~(in this case the mixture), the ranges of the parents, and the incoming $\pi$ messages, whose type is a \texttt{Dist}. (This also illustrates another abstraction of Scruff; rather than represent the $\pi$ messages as a vector of floats, as it would be in most other PP systems, we let it be any \texttt{Dist}. We will discuss this point more in the next section).
The implementation of \texttt{ComputePi} is as follows:

\begin{verbatim}
1 @op_impl function ComputePi{Mixture}((range, parranges, incoming_pis))
2    function f(i)
3         cp = compute_pi(sf.components[i], range, parranges, incoming_pis)
4         sf.probabilities[i] .* [cpdf(cp, (), x) for x in range]
5     end
6     scaled = [f(i) for i in 1:length(sf.components)]
7     result = sum(scaled)
8     return Cat(range, normalize(result))
9 end
\end{verbatim}

Lines 2 to 5 defined a function that is performed for every index \texttt{i} from 1 to the number of components. (Julia array indices start with 1). For each component, the function delegates to the \texttt{compute\_pi} implementation of the component, with the same values of the arguments. (Note the automatic transformation of case from Julia's standard type case (\texttt{ComputePi}) to function case (\texttt{compute\_pi})). In Scruff's abstract implementation, this can return any kind of \texttt{Dist}. Therefore, the operation \texttt{cpdf} (which stands for conditional probability density function) is used to get the probability of each element in \texttt{range}. This is the operation through which any \texttt{Dist} can be converted to the usual vector of floats representation. The function \texttt{f} then uses Julia broadcasting to multiply each of these probabilities with \texttt{sf.probabilities[i]}, which is the mixture probability for the corresponding component. Once this function has been defined, most of the work is done. The function is invoked for each component index in line 6, and the vectors for each component are summed in line 7 and normalized in line 8. Since a \texttt{Dist} must be returned, this result is packaged into a \texttt{Cat} (for categorical).

\texttt{Mixture} is perhaps a simple and fairly standard example, but before we move on to more sophisticated composition operators, we show how Scruff uses the \texttt{Mixture} abstraction to represent an interesting class of \sfuncs~and to automatically derive \opimpls~for them.
A \texttt{Separable} \sfunc~\cite{pfeffer2001sufficiency} is a representation of a conditional probability table for $X$ given parents $U_1,\ldots,U_n$ with special structure: $P(X \mid U_1,\ldots,U_n)$ decomposes into $\sum_{i=1}^n P_i(X\mid U_i)$. \texttt{Separable} \sfuncs~are desirable for both representational and computational reasons. The representation of \texttt{Separable} uses a number of parameters that is linear in $n$, rather than exponential in $n$ for the usual representation. Furthermore, the cost of performing BP operations such as \texttt{ComputePi} is also linear in $n$ for \texttt{Separable}~\sfuncs, rather than exponential. This fact was actually not known to us until we developed an explicit \texttt{Separable} implementation rather than relying on standard conditional probability table implementations.

The additive decomposition in the definition of \texttt{Separable} \sfuncs~suggests a mixture, but each of the $P_i$ has a different parent $U_i$, violating the standard assumptions of a mixture. To fix this, Scruff provides an \texttt{Extend} \sfunc~that takes an \sfunc~with one parent and converting it to an \sfunc~with many parents, while ignoring all the new parents. By applying \texttt{Extend} to the individual $P_i$, we align them to the same set of parents and can then represesnt the separable \sfunc~as a mixture. Once this is done, no operations need to be implemented for \texttt{Separable}; they are all derived automatically from \texttt{Mixture}. In fact, the derived version of \texttt{ComputePi} implements the efficient linear time algorithm we discovered. In fact, once we realized that \texttt{Separable} can be represented as a \texttt{Mixture}, the algorithm we discovered became obvious, because the implementation of \texttt{ComputePi} for \texttt{Mixture} is straightforward. This shows the potential of Scruff's abstract approach to provide useful and efficient new implementations.

\subsection{Conditionals}

\begin{figure}[h]
	\centerline{\includegraphics[scale=0.3]{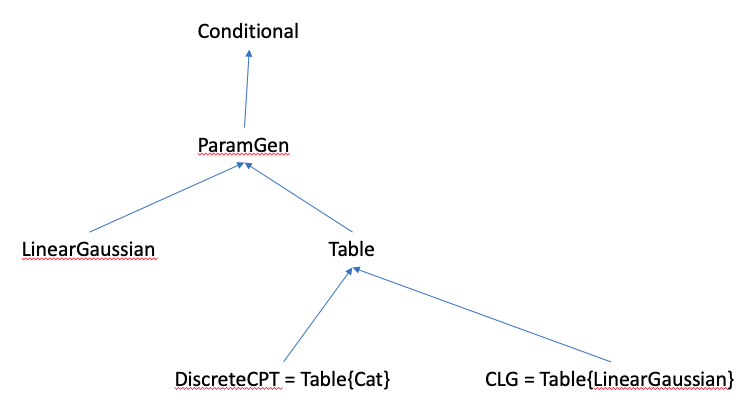}}
	\caption{Hierarchy of conditional probability relationships}
	\label{fig:conditional}
\end{figure}

Scruff provides a rich class hierarchy, rooted in the \texttt{Conditional} \sfunc, to represent different kinds of conditional probability relationships.
To orient the reader through this subsection, Figure~\ref{fig:conditional} shows a hierarchy of the \sfuncs~we will discuss.

At the root of the hierarchy, \texttt{Conditional} captures the general notion of probabilistic dependence, without minimal assumptions about how that dependence is represented. The definition of \texttt{Conditional} is as follows:

\begin{verbatim}
abstract type Conditional{I <: Tuple, J <: Tuple, K <: Tuple, O, 
                                       Q, P, S <: SFunc{J, O, Q}} 
  <: SFunc{K, O, P}
end
\end{verbatim}
The plethora of type parameters indicates that there is a lot to unpack here.
A \texttt{Conditional} has two sets of parents, known is the $I$ parents and the $J$ parents.
The type parameters \texttt{I} and \texttt{J} represent the tuple types for these two sets of parents. 
The type parameter \texttt{K} represents the concatenation of \texttt{I} and \texttt{J} into a single tuple of parents. (As far as we know, Julia cannot express this concatenation without having to introduce the additional type parameter.)The type \texttt{O} is the output type of the \texttt{Cconditional}.

The $I$ and $J$ parent sets serve two different roles. The \texttt{Conditional} has an embedded \sfunc~type \texttt{S}, which is an \texttt{SFunc\{J, O, Q\}}. That is, its input is a $J$ parent set and its output is \texttt{O}, which is the output of the \texttt{Conditional}. \texttt{S} also has parameters of type \texttt{Q}. Thus, the role of the $J$ parent set is to serve as parents of another \sfunc. The role of the $I$ parent set is to determine which other \sfunc~to use, so long which must be of type \texttt{S}. If we think of \sfuncs~as generative models, we can imagine a generative process of looking at the $I$ parents to choose an \sfunc~$s$, and then using the generative process of $s$, given the $J$ parents, to generate an output $o$. However, we caution that \sfuncs~are not necessarily generative models, but the idea of using the $I$ parents to choose the relationship and the $J$ parents to participate in the relationship holds in general.

The \texttt{Conditional} as a whole is an $\texttt{SFunc\{K, O, P\}}$. We have already seen that \texttt{K} is the concatenation of the $I$ and $J$ parent type tuples, while $O$, the output of the embedded \sfunc~type \texttt{S} is also the output type of the \texttt{Conditional}. \texttt{P} is the type of parameters of the \texttt{Conditional}, which are different from the parameter type \texttt{Q} of the selected \sfuncs. 

It is surprising how many \opimpls~can be developed even at this very  abstract level. These implementations capture the essence of the process of using a set of parents (the $I$ parents) to choose a probabilistic relationship that may depend on other parents (the $J$ parents). Operations implemented for \texttt{Conditional} include computing support, drawing a sample from the conditional probability distribution, computing the log conditional probability density, creating a tabular factor to represent the conditional probability distribution, as well as the BP operations \texttt{ComputePi} and \texttt{SendLambda}. All these implementations rely on a method \texttt{gensf} that takes a tuple of values of $I$ parents and returns the \sfunc~of type \texttt{S} selected.

Unfortunately, the implementations of these operators are either trivial (as in the case of \texttt{Sample} or \texttt{Logcpdf}) or too long to show here. We will show the code for \texttt{Sample} to illustrate the easy case and show pseudocode for some of the other cases.
Here is the implementation of \texttt{Sample}:

\begin{verbatim}
@op_impl function Sample{Conditional}(parvals)
    (ivals, jvals) = split_pars(sf, parvals)
    sfg = gensf(sf,ivals)
    return sample(sfg, jvals)
end
\end{verbatim}

\texttt{split\_pars} is a utility function that decomposes a tuple of values into $I$ and $J$ tuples.
This code illustrates why \texttt{Sample} is so easy to implement; it simply selects the \sfunc~to use for the given $I$ parent and then delegates sampling to it.
In general, however, the implementation needs to reason about and integrate different possible choices of the $I$ parents.
The more complex \opimpls~demonstrate that this reasoning is general. For example, here is pseudocode for \texttt{ComputePi}:

\texttt{ComputePi(sf::Conditional, range, parranges, incoming\_pis)}
\begin{enumerate}
\item Split \texttt{parranges} and \texttt{incoming\_pis} into $I$ and $J$ components, called (\texttt{iranges},\texttt{jranges}) and (\texttt{ipis},\texttt{jpis}).
\item Compute the Cartesian product of \texttt{iranges}.
\item For each tuple $i$ in the Cartesian product:
\begin{enumerate}
    \item $R \leftarrow \texttt{zeros(length(range))}$.
	\item $p_i \leftarrow \prod_{k=1}^n \texttt{ipi}_k(i_k)$.
	\item $s \leftarrow \texttt{gensf}(\texttt{sf},i)$.
	\item $p_j = \texttt{ComputePi}(s,\texttt{range},\texttt{jranges}, \texttt{jpi})$.
	\item $R \leftarrow R + p_i~\mbox{.*}~p_j$ \# element-wise product
\end{enumerate}
\item Return $R$
\end{enumerate}

The important thing about this implementation is not just the fact that it can be implemented but what it says about the general process being described. In short, it sums, over all the possible cases of $I$, the probability of that case times the $\pi$ value computed for the selected \sfunc~for that case. This is the essence of this operation for conditional relationships and it has wide applicability. All the other operations on \texttt{Conditional} are implemented along similar lines.

Before we get to the applications, we introduce a useful special case of \texttt{Conditional}, known as \texttt{ParamGen}. \texttt{ParamGen} stands for Parameter Generator, which is the role of the $I$ parents, who select the \sfunc~$s$ by setting the parameters of a fixed embedded \sfunc. This allows a particularly easy implementation of \texttt{gensf} using a function \texttt{gen\_params}, which takes a value of the $I$ parents and returns the parameters for the embedded \sfunc. The significance of \texttt{ParamGen} is that it enables the implementation of operations related to the Expectation-Maximization (EM) algorithm, including computing expected sufficient statistics and maximizing the parameters given expected sufficient statistics. As a result, \sfuncs~that derive from \texttt{ParamGen} are able to participate in EM learning processes by parameter estimation.

One application of \texttt{Conditional} and \texttt{ParamGen} is the linear Gaussian model. A linear Gaussian represents a conditional distribution of one real-valued child on a number of real-valued parents. Specifically, the child is Gaussian (i.e, normally) distributed with a mean that is a linear function of the values of the parents.

Scruff implements \texttt{LinearGaussian} in a straightforward way by deriving from \texttt{ParamGen}. 
In this case, all the parents are $I$ parents and there are no $J$ parents.
The embedded \sfunc~is a \texttt{Normal}.
The \texttt{gen\_params} function takes the values of the parents and returns the output of the linear function and sets the mean of the \texttt{Normal} appropriately with fixed variance.

Another application is the \texttt{Table}, which is actually a general kind of \texttt{ParamGen} where the $I$ parents select the appropriate \sfunc~$s$ by indexing into a table of parameters. Again, the implementation is straightforward. The $I$ parents are used for table lookup, while the $J$ parents are the parents of the embedded \sfunc. The \texttt{gen\_params} function simply performs the table lookup.

Once we have \texttt{Table}, we can define a variety of different representations of a conditional probability table. For example, \texttt{DiscreteCPT} is a traditional conditional probability table as it appears in standard discrete Bayesian networks (BNs). In this case, the embedded \sfunc~is a \texttt{Cat}, i.e., an explicit probability distribution over a discrete set of values.

A richer representation is the \texttt{CLG}, which stands for \emph{conditional linear Gaussian}. 
In the BN literature, a CLG is viewed as a representation for a continuous child with a mix of continuous and discrete parents~\cite{lauritzen1992propagation}. The discrete parents determine which specific linear Gaussian \sfunc~to use for the relationship between the continuous parents and the child. In Scruff, a \texttt{CLG} is simply a \texttt{Table} in which the embedded \sfunc~is a \texttt{LinearGaussian}. 

Although algorithms for inference with BNs that include CLGs are well-known and implemented in commercial-grade systems like HUGIN~\cite{andersen1989hugin}, they are not easy to implement in general and non-commercial systems often do not support them. In Scruff, the \opimpls~are derived automatically from the operations on \texttt{Normal}, \texttt{ParamGen}, and \texttt{Conditional}, so CLGs can participate in any algorithm that uses these operations, including VE, BP, and EM, in addition to easier-to-implement algorithms like importance sampling and Markov chain Monte Carlo. Furthermore, they can be combined with a far wider variety of representations than in traditional software.

\subsection{Other Composition \sfuncs}
	
While the \texttt{Conditional} family of \sfuncs~captures a general class of probabilistic dependencies, Scruff provides a number of other composition \sfuncs, besides \texttt{Mixture} which we presented earlier. These include \texttt{Det} for deterministic relationships; \texttt{Switch} and \texttt{If} to represent conditional execution; and \texttt{Apply}, which is a type of higher order \sfunc. We discuss each of these in turn.

\texttt{Det} very simply represents a deterministic function from the inputs to the output.
This is extremely useful, for example, to include physics models in Scruff programs.
We have used this technique to integrate physical models of electrical components with generative probabilistic models of system behavior for health monitoring applications.

 The implementation of \texttt{Sample} is straightforward: simply apply the given function to the inputs and return the result. Many of the other operations, such as \texttt{ComputePi}, are also straightforward; if we have a distribution over the inputs, we can easily get a distribution over the output. 

However, operations that invert the \sfunc~by attempting to infer properties of inputs given observations on outputs are more difficult, for multiple reasons. First, it is not in general tractable to invert a function to determine what inputs might have led to an observed output. Second, many candidate output values according to the \sfunc~type might not correspond to any given input, so they have probability zero. This can be a source of significant error.

While Scruff does not have a complete solution to this fundamentally difficult problem, the approach of allowing multiple implementations of operations provides a developer the freedom to implement partial solutions that work well in some cases, enabling these \sfuncs~to be used in situations that might not otherwise have been possible. For example, one implementation Scruff provides of the \texttt{SendLambda} operation uses interpolation. Given ranges of values of the parents, potential values of the child are generated by calling the given function. Observed values of the child are then compared to the generated values by interpolation, leading to evidential $\lambda$ messages being sent to the parents. Although this technique will not work in general, it seems reasonable for some physical models, and works well in our applications. It's also important to note that \opimpls do not need to be exact, even if you wish to employ exact inference - any approximate inference algorithm can be converted to an asymptotically exact one by using it as a proposal distribution for importance sampling or similar algorithm. In this way even very rough approximations can result in major improvements over e.g. sample-based methods.

\texttt{Switch} is a simple \sfunc~that derives from \texttt{Det} to represent situations in which one of a set of \sfuncs~is chosen based on the state of a switch. \texttt{Switch} can be compared to \texttt{Mixture}. In \texttt{Mixture}, the choice of \sfunc~is stochastic based on a probability distribution, while in \texttt{Switch}, the choice is deterministic based on another parent. 

These connections suggest alternative implementations.
For example, \texttt{Mixture} could be implemented by a \texttt{Switch} in which the switch is implemented as a \texttt{Cat}. It would be interesting to explore whether \opimpls~for this more general representation will recover the efficient implementations for \texttt{Mixture} and \texttt{Separable}. Finally, we remark that using \texttt{Switch}, we can also implement a simple conditional \texttt{If} where the switch is Boolean and there are two choices.
	
Finally, \texttt{Apply\{J,O,P\}} is a higher-order \sfunc~that has two inputs. The first input is an \texttt{SFunc}\{\texttt{J},\texttt{O},\texttt{P}\}, which takes the second input of type \texttt{J} as parent. \texttt{Apply} implements the concept of functional uncertainty in PP, which is uncertainty over the identity of the function to use in a function application. 
This can be seen by considering a network with three \sfuncs:
\begin{enumerate}
	\item A \texttt{Dist\{S,Q\}} where $S$ is \texttt{SFunc\{J,O,P\}}, with no parents.
	\item A \texttt{Dist\{J,R\}} where $J$ is any type, with no parents.
	\item An \texttt{Apply\{J,O,P\}} with the first two \sfuncs~as parents.
\end{enumerate}

Again, there are intersting connections to explore that may be fruitful.
\texttt{Conditional} might be implemented using \texttt{Apply},
in which the \texttt{gensf} function is reified in a \texttt{Det} that takes the $I$ parents as arguments and returns the selected \sfunc.
We currently only have a small set of operations implemented for \texttt{Apply}; this is work in progress, and it would be interesting to see whether it is possible to reimplement \texttt{Conditional} using this more general mechanism and recover the \opimpls.

\section{Algorithms}
\label{sec:algorithms}

One of our main claims is that a small set of well-chosen operations can support a variety of algorithms in different representational frameworks. This has two main benefits. First, it enables us to generalize implementations of algorithms across paradigms, reducing development effort. Second, and more significantly, it enables us to combine different paradigms in a single application, using networks that have different kinds of \sfuncs~in them. 

In this section, we show how several algorithm families can be treated in this way. We discuss elimination algorithms, message passing algorithms, and generate and score algorithms. However, we do not claim that all algorithms are amenable to this approach. For example, Markov chain Monte Carlo (MCMC) algorithms, widely used in probabilistic programming, are as far as we know specific to probabilistic models. Although they can be readily incorporated into Scruff, the two benefits above are not achieved for these algorithms.

In this section, we build on the work of Dechter~\cite{dechter1999bucket, dechter2019reasoning}, who has unified different AI frameworks under the umbrella of elimination and conditioning algorithms. We go beyond this work in two ways. We consider a wider variety of frameworks, including NNs and ordinary differential equations (ODEs) for inclusion in this approach. We also present a general software paradigm for implementing these algorithms that is flexible because it allows the algorithm developer to combine different operations as desired.

\subsection{Elimination Algorithms}

Elimination algorithms, of which VE~\cite{zhang1994simple, dechter1998bucket} is the canonical example, work by systematically removing \sfuncs~from a network until only a target set of \sfuncs~remains, with information computed about those target \sfuncs. 
In general, these algorithms formulate a computation as a sum-of-products expression over factors and use the commutative and distributive laws to manipulate this expression to reduce computation.
Expressed in this way, several authors have shown that elimination algorithms can be applied to any semiring~\cite{goodman1999semiring, pfeffer2007design, eisner2004dyna}.

For example, for standard BN inference, factors are tables of real numbers, sum is the usual sum on reals, and product is the usual product on reals. Computing the most likely explanation is similar, except that sum is a maximization function. For logical systems, sum is disjunction and product is conjunction. For EM learning, sum and product operations also process sufficient statistics, which often also form a semiring. For decision-theoretic problems, sum and product process utilities, which are also a semiring.

To implement elimination algorithms, Scruff uses a \texttt{Semiring} class hierarchy, where each type of \texttt{Semiring} has appropriate \texttt{sum} and \texttt{product} functions.
Each of the above algorithms will use a different type of \texttt{Semiring}; although probabilistic inference and most likely representation have the same factor representation, they use different \texttt{sum} functions.
To implement an elimination algorithm, a \texttt{MakeFactors} operation, which creates the factor for the \sfunc~over the appropriate \texttt{Semiring}, needs to be implemented for the relevant \sfuncs.
\texttt{MakeFactors} takes the \texttt{Semiring} as an argument, enabling the same \sfunc to be used in different elimination algorithms.

Once the \texttt{Semiring} type and corresponding \texttt{MakeFactors} operation has been implemented, the VE algorithm is obtained. Furthermore, there are many variants of elimination algorithms, such as the junction tree algorithm~\cite{lauritzen1988local} used in commercial-grade BN implementations and the mini-buckets approximation scheme~\cite{dechter2003mini}. If these algorithms are implemented using the \texttt{MakeFactors} operation, they are obtained automatically for all the different frameworks.

Scruff's expression of all these algorithms in a common framework is not novel; other PP systems (e.g. IBAL~\cite{pfeffer2007design}) have used semirings to implement a variety of elimination algorithms.
However, implementing these methods within a standard flexible framework leads to 
surprising new possibilities.
The \texttt{product} function each take two values as arguments.
In Julia, which uses multiple dispatch, we are able to define different methods for these functions that take different \texttt{Semiring} types as arguments.
For example, we might provide an implementation of \texttt{product} that takes a logical value and a probabilistic value and zeros out the probabilistic value when the corresponding logical value is \texttt{false}. 
We would then obtain elimination algorithms that work jointly on networks including both probabilistic and logical \sfuncs. 
An alternative way to implement this idea is to use the standard probabilistic semiring also to implement logical elimination algorithms, but this would be unnatural for implementers of the logical algorithms.
Furthermore, integrating the framework in the way we have described allows for alternative implementation; for example, we might combine the logical and probabilistic entries by treating the probabilistic entry as a Boolean that is \texttt{true} if the value is non-zero.

\subsection{Message Passing Algorithms}

The name "propagation" is common to many algorithms used in AI, including constraint propagation, belief propagation, and NN backpropagation. This sharing of name is not a coincidence. All these algorithms are based on message passing on graphs. In general, message passing on directed graphs includes three kinds of operations: "forward" message computations, "backward" message computations, and local updates. However, from now on we will avoid using the terms "forward" and "backward" as these have opposite meanings in different communities. In the probabilistic reasoning community, "forward" is the generative direction from causes to effects, while "backward" is the direction of inference from observations on effects to beliefs about causes. In feed-forward NNs, "forward" is the direction of inference, i.e. predicting unknown values (e.g. classifications) from observations on effects (e.g. images), while "backward" is the direction of inferring gradients on weights in the network from errors in the prediction. Instead of "forward" and "backward", we will follow Pearl's terminology and use the $\pi$ direction for the direction from causes to effects and the $\lambda$ direction for observations on effects to inferences or predictions about causes.

Generative probabilistic models and NNs have complementary strengths and weaknesses. Generative models are generally good at computing in the $\pi$ direction, which requires sampling or computing values. These generative models have predominated in PP, which is one reason why sampling algorithms have been so prevalent. While reasoning in the $\lambda$ direction can be achieved, and much effort has been expended on making it efficient, in general it requires inverting the \sfunc~and can be very expensive.
Meanwhile, NNs are very good at reasoning in the $\lambda$ direction, where prediction is a linear time operation in the size of the network. However, it is extremely difficult to infer inputs given outputs in the $\pi$ direction.
ODEs can generally be used well in both directions.

These facts have led researchers to develop bidirectional models that use a generative probabilistic model in the $\pi$ direction and a NN in the $\lambda$ direction.
The intention is that the that the two models express (approximately) the same relationship between inputs (in the Scruff sense of inputs to an \sfunc) and outputs.
Inference compilation techniques~\cite{le2017inference} implement this idea by using the generative model to generate a training set to train a NN that can be used in the $\lambda$ direction.
Variational autoencoders~\cite{kingma2014stochastic} are another implementation of this idea.


Scruff implements message-passing in a general way through operations that reason in the $\pi$ direction, $\lambda$ direction, or locally within an \sfunc. For example, we have already shown \texttt{ComputePi}, which produces a representation of the prior knowledge about a variable, and \texttt{SendLambda}, which sends a $\lambda$ message to a particular parent.
An example of a local update operation is \texttt{ComputeBel}, which combines prior and evidential knowledge to form beliefs about the variable. In Bayesian reasoning, this is the process of multiplying prior probabilities with likelihoods. In other frameworks, \texttt{ComputeBel} might be implemented lazily, using a function that gets the appropriate belief when desired. In Section~\ref{sec:policies}, we show some examples of alternative implementations along with policies to choose between them.

\begin{figure}[h]
	\centerline{\includegraphics[scale=0.3]{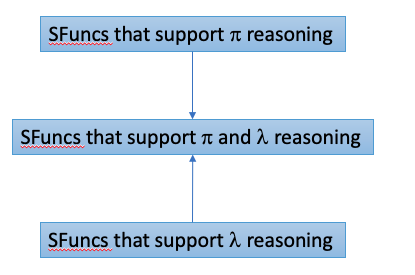}}
	\caption{Structure of Scruff networks used with message passing}
	\label{fig:message}
\end{figure}

Figure~\ref{fig:message} shows the general structure of Scruff networks that are used in message passing. The top layer consists of \sfuncs~that can support $\pi$ operations but not $\lambda$ operations. These include, for example, \sfuncs~that are hard to invert. As we have discussed, \texttt{Det} \sfuncs~can be hard to invert. Although we have implemented an approximation method using interpolation that enables them to be used in a more general way, a safe way to use them is to include them in the top layer of the network. The drawback is that it is not impossible to send evidential messages to their parents. Despite this, they can still have interesting parent-child relationships and can be used to represent rich priors, including from physical models.

The bottom layer consists of \sfuncs~that support $\lambda$ operations but not $\pi$ operations. These are most typically used to interpret sensor observations, for example with NNs. Since the sensors are observed, the inability to propagate priors to them is not a significant drawback.

The middle layer consists of \sfuncs~that support both $\pi$ and $\lambda$ operations and can participate fully in the message passing process.
This includes bidirectional \sfuncs~as well as most of the examples of probabilistic \sfuncs~we discussed in Section~\ref{sec:sfuncs}.
Note that the other two layers can also include these types of \sfuncs.
However, they will not be able to participate fully in message passing, because they are ancestors of \sfuncs~that do not support $\lambda$ operations or descendants of \sfuncs~that do not support $\pi$ operations.

The benefit of this approach is that it enables \sfunc~developers to support the operations they are able to support without worrying about completeness of implementation, while algorithms can be flexible about using the \sfuncs~in the appropriate way.
For example, a BP algorithm can begin with an analysis that determines on which variables it can perform $\pi$ and $\lambda$ operations, and then proceed to do as much work as possible, obtaining beliefs about as many variables as possible.

To this point, we have focused on inference but not learning, which is of course vital in many modeling frameworks.
In NNs, stochastic gradient descent is used to train the weights of the network.
Although EM is usually used to learn the parameters of Bayesian networks in practice, it is well-known that the parameters can be learned by gradient descent as well~\cite{russell1995local}.
In an earlier paper, we showed (citation surpressed), using automatic differentiation techniques, that gradient descent computations
for Bayesian networks can be achieved as a byproduct of the BP process,
using message passing.
Scruff's support for gradient-computation message passing algorithms is very similar to its support for inference. 
The major difference is in the direction of the messages.
In NNs, inference is used in the $\lambda$ direction to predict hidden state, while gradient messages proceed in the $\pi$ direction to process the errors.
For BNs, predictions of outputs is done in the $\pi$ direction, and then gradient messages are then processed in the $\lambda$ direction.

\begin{figure}[h]
	\centerline{\includegraphics[scale=0.3]{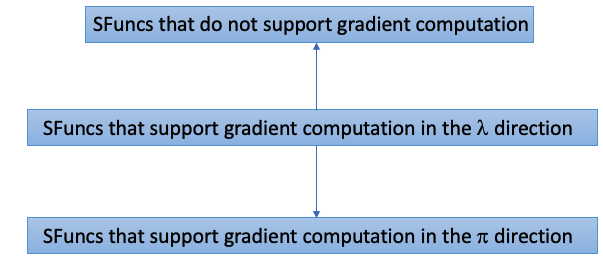}}
	\caption{Structure of Scruff networks used with gradient descent}
	\label{fig:gradient}
\end{figure}

Figure~\ref{fig:gradient} shows the structure of networks used in gradient descent, similarly to Figure~\ref{fig:message}. Fortunately, many \sfuncs~will play the same role in both structures.
For example, NNs will appear in the bottom layer, while deterministic functions, which are generally not learnable, might be in the top layer.

\subsection{Generate and Score Algorithms}

Generate and test is a standard algorithmic framework used to search for an value that satisfies certain properties. It requires a generator of value and a way to determine whether an value satisfies the required properties. Values are repeatedly generated by the generator and then tested; only values that pass the test are preserved. In probabilistic inference, this is the basis of the rejection sampling algorithm.

Generate and test can be generalized to generate and score algorithms. Instead of a Boolean test whether an item satisfies a property, it is scored with a number, and items generated are weighted by their score. In probabilistic inference, generate and score is implemented by importance sampling.
Importance sampling is also the basis for sequential Monte Carlo algorithms, also known as particle filtering, widely used to reason about dynamic probabilistic models.

These ideas are so simple as to be ubiquitous. Scruff provides a very general implementation of these ideas using a \texttt{Sample($s$)} operation to generate a value from the \sfunc~$s$ and a \texttt{GetScore($c,x$)} operation to score the value $x$ with the \texttt{Score} $c$. As we have discussed, the \texttt{Score} family of \sfuncs~is flexible and can include, for example, both explicit vectors and functional computations. Although the ideas here are simple, the flexibility is useful.

Things get interesting, however, when we combine these generate and score algorithms with other kinds of algorithms. A drawback of generate and score algorithms is that it can be difficult and expensive to generate values that have reasonably high scores and are representative of the values of interest. To mitigate this problem, researchers have developed lookahead sampling techniques~\cite{dearden2002particle, ahmed2012decentralized} that use the observations to guide the generation of samples. A simple implementation of lookahead sampling is likelihood weighting~\cite{sato1997prism, shwe1991empirical}, which only looks ahead one step.
In probabilistic programming, researchers have used various strategies to look ahead in importance sampling~\cite{pfeffer2007general, nori2014r2}.

Using Scruff, we get a highly general version of lookahead sampling with little effort by combining it with message passing. If we consider again the structure shown in Figure~\ref{fig:message}, we see that $\lambda$ messages can be propagated all the way up except into the top layer.
$\lambda$ messages are in fact represented as instances of $\texttt{Score}$.
Therefore, they can be used in a generate and score algorithm to score the values generated by the top layer. 
Dechter has developed in depth similar ideas combining conditioning with elimination~\cite{mateescu2012relationship, dechter2019reasoning}.
Scruff's approach demonstrates how a simple set of operations can be used to easily create implementations of algorithms that have hitherto required significant research.

\section{Policies}
\label{sec:policies}

A policy in Scruff is a program that, upon invocation of an operation, selects a compatible operation implemetation to employ as well as values for any available hyperparameters. We refer to this as a policy with inspiration from Markov Decision Processes (MDP);
a policy determines the choice of \opimpl~at every point.
The act of selecting \opimpls~over the course of algorithm execution with the goal of producing an accurate, fast result essentially defines an MDP. Automatically deducing a performant set of operation choices is thus equivalent to a policy optimization problem.
However, methods for performing these optimizations are left for future work. For now, we allow the programmer to explicitly implement policies, but have also provided the framework for automatic optimization in future.

Policies are programs that take as input the inputs, outputs and SFuncs associated with each operation call during the execution of an algorithm. Then, when an operation is invoked, the policy can use any computational rule to produce a valid OpImpl for the invoked operation and \sfunc.
If no policy is specified, the default is to rely on Julia's standard methods using multiple dispatch.
Scruff provides support functions for finding all the candidate \opimpls~that could be used for a given call.

Policies can make use of performance characteristics of \opimpls, which are implementer-specified descriptions of expected performance of the \opimpl.
These are implemented using the \texttt{@op\_perf} macro, which defines an operation performance chartacteristic or \opperf. The macro takes an \opimpl~type and the arguments to the \opimpl~and returns a standard performance measure.
For example, the \texttt{runtime} performance measure estimates, naturally, the running time of the \opimpl~on the given arguments.
The compositional nature of \sfuncs~also enables composition of \opperf~implementations. For example, the runtime of calling \texttt{ComputePi} for a \texttt{Mixture} is approximately the sum of the runtimes for the components.

One useful performance measure in Scruff is \texttt{support\_quality}. 
Scruff relies heavily on the ability to compute the support of \sfuncs~for a variety of algorithms.
This support computation can have different characteristics, which are useful for algorithms to know.
\begin{itemize}
\item For some \sfuncs, particularly those with a small discrete range, this is straightforward and can be done completely. This is important for algorithms that want to provide exact answers.
\item For others, such as continuous \sfuncs~like \texttt{Normal}, the range cannot be enumerated fully.
Nevertheless, Scruff provides operations for computing the support of \sfuncs~like \texttt{Normal} incrementally, producing increasingly rich support sets on each iteration, with the support set after each iteration being a superset of the previous iteration.
One of the algorithms in Scruff is a lazy algorithm that iteratively refines a network until the desired accuracy is achieved, which makes use of this incremental support computation.
The iterative improvement quality enables this algorithm to guarantee improvement of the answers computed on each iteration.
\item 
For other \sfuncs, support computation is neither complete nor incremental, but done on a best effort basis. Although these provide no guarantees, they can still be useful in approximate algorithms.
\end{itemize}

One way to deal with these different kinds of guarantees about the support computation would be to use three different operations for complete, incremental, and best effort support. However, this would be both inconvenient and misleading, since they all intend to compute exactly or approximately the same mathematical definition, which is the set of values that have positive probability. The difference is only in the success they can offer.
Therefore, we use a single \texttt{support} operation for all these cases and express the different guarantees using \texttt{support\_quality}. An implementation of \texttt{support} will declare its guarantees using an \opperf. For compositional \sfuncs, the implementation of the \opperf~will recurse to the arguments. For example, the \texttt{support\_quality} for \texttt{Mixture} will be the minimum \texttt{support\_quality} of the components, where the ranks are complete, incremental, and best effort.

Another useful performance characteristic is laziness.
Some \opimpls~will compute values eagerly while others will defer them until later.
For example, the \texttt{compute\_bel} operation takes a \texttt{Dist}, representing information from the inputs coming in the $\pi$ direction, and a \texttt{Score}, representing information from the outputs coming in the $\lambda$ direction, and computes a belief, i.e. a distribution over the value of the variable considering all the information, which is also implemented as a \texttt{Score}.
Mathematically, the belief is the element-wise product of the \texttt{Dist} and the \texttt{Score}.
While simple implementations can be created for specific types of \texttt{Dist} and \texttt{Score}, there are generic implementations that work for any type of \texttt{Dist} and \texttt{Score}. Here are two implementations, one eager and one lazy. The eager implementation needs to immediately produce a score, so it needs to sample values of the variable. The lazy implementation returns a function that can be used later to compute the belief and wraps it in a \texttt{FunctionalScore}.

\begin{verbatim}
@op_impl begin
    struct EagerComputeBel <: ComputeBel{SFunc} 
        num_samples :: Int # hyperparameter
        EagerComputeBel() = new(100) # default value of hyperparameter
    end
    function compute_bel(d::Dist, s::Score)
        values = Vector{I}()
        scores = Vector{Float32}()
        for i in 1:num_samples
            x = sample(d)
            push!(values, x)
            push!(scores, score(s,x))
        end
        return SoftScore(values, scores)
    end
end

@op_impl begin
    struct LazyComputeBel <: CompiuteBel{Sfunc}
    function compute_bel(d::Dist, s::Score)
        f(x) = density(u,x) * score(s,x)
        return FunctionalScore(f)
    end
end
\end{verbatim}

The laziness characteristic of the two \opimpls~are declared using \texttt{is\_lazy}. Specifically,
the following \opperf~declarations are used:
\begin{verbatim}
@op_perf is_lazy(EagerComputeBel) = false
@op_perf is_lazy(LazyComputeBel) = true
\end{verbatim}

Algorithms that wish to make a preference, say, for lazy inference, can use \texttt{is\_lazy} in their policy implementations.
Given a choice between multiple \opimpls, they can prefer the lazy ones, perhaps along with other criteria.

\section{Future Prospects}
\label{sec:future}

Scruff has been under development for several years. Although work is ongoing, we feel that much has been achieved and the time is right for publication.
We are planning an open source release in the next few months
(the release will be delayed until after the POPL reviewing period). 

We would like to highlight two potential directions for future work.
The first is meta-reasoning.
We have already hinted at meta-reasoning in the discussion of policies.
Determining a policy for choosing \opimpls~is one form of meta-reasoning.
Our \opperf~framework provides an initial basis for this.
More interesting would be to view \opimpls, whose performance is uncertain,
as themselves being \sfuncs.
\opperfs~would then become \opimpls~for \opimpls~as \sfuncs.
This approach could potentially enable us to create systems that reason about their environment and their internal operation in an integrated way.

A second potential direction is to AI planning systems. 
Planning has been integrated into PP using the planning as inference paradigm~\cite{botvinick2012planning},
where planning is accomplished by inferring the actions that are likely to lead to desirable states.
However, it is possible that a wider variety of AI planning algorithms could also be integrated into the Scruff framework.
Many planning algorithms can be viewed as message passing using forward or backward reasoning over a set of planning operators.
Achieving this unification could enable new forms of AI agent design that do joint inference and planning.

\bibliographystyle{apalike}
\bibliography{implicit}



\end{document}